\documentclass[lettersize, journal]{IEEEtran} 

\usepackage[utf8]{inputenc}
\usepackage{glossaries}
\usepackage{comment}
\usepackage{float}
\usepackage{amsmath}
\usepackage{amssymb}
\usepackage{hyperref}
\usepackage{amsthm}

\newcommand{\desiredpositions}{\mathbf{p}^\mathrm{d}}
\newcommand{\positions}{\mathbf{p}}
\newcommand{\xposition}{p_\mathrm{x}} 
\newcommand{\yposition}{p_\mathrm{y}}
\newcommand{\state}{\mathbf{s}}
\newcommand{\orientation}{\phi}
\newcommand{\turningrate}{\omega}
\newcommand{\nvelocity}{v} 
\DeclareMathOperator{\nacceleration}{a} 
\DeclareMathOperator{\control}{\mathbf{u}} 

\newcommand{\adaptivewaypoints}{rule-adaptive waypoints}

\newcommand{\waypointadaption}{rule-adaptive waypoint engine}

\usepackage[doi=false,isbn=false,url=false,eprint=false]{biblatex} 
\usepackage{listings}
\usepackage{color}
\definecolor{tumblue}{rgb}{0, 0.4, 0.74}
\usepackage[scaled=.7]{beramono}

\usepackage{tikz,pgfplots}
\usepackage{textcomp}
\usepackage{makecell}
\usepackage{fontawesome}
\usepackage{booktabs}
\usepackage{enumitem}

\usepackage{siunitx}
\usepackage{graphicx}
\usepackage{epstopdf}
\usepackage{calc}
\usetikzlibrary{positioning,chains}
\usetikzlibrary{shapes.geometric}
\usetikzlibrary{calc}
\usetikzlibrary{trees}
\usetikzlibrary{decorations.pathreplacing,calc,tikzmark}
\usetikzlibrary{backgrounds, fit}
\usetikzlibrary{positioning,arrows,petri,shapes.misc}
\usetikzlibrary{arrows.meta, bending, calc, fit, shapes, patterns}
\usepackage{algorithm}
\usepackage{algpseudocode}

\definecolor{TUMBlue}{RGB}{0, 101, 189}
\definecolor{TUMOrange}{RGB}{227, 114, 34}
\definecolor{TUMGreen}{RGB}{162, 173,   0}
\definecolor{TUMGray}{RGB}{51,  51,  51}
\usepackage{pifont}

\usepackage[nohyperlinks, nolist]{acronym}
\input{figures/plotstyle}
\makeatletter
\let\MYcaption\@makecaption
\makeatother

\usepackage[font=footnotesize]{subcaption}

\makeatletter
\let\@makecaption\MYcaption
\makeatother

\begin{document}
\begin{acronym}
	\acro{colregs}[COLREGS]{Convention on the International Regulations for Preventing Collisions at Sea}
	\acro{rl}[RL]{Reinforcement learning}
	\acro{mpc}[MPC]{model predictive control}
    \acro{ism}[ISM]{Intelligent Sailing Model}
     \acro{ais}[AIS]{Automatic Identification System}
\end{acronym}

\title{Intelligent Sailing Model for Open Sea Navigation}

\author{Hanna Krasowski$^{*, 1}$, Stefan Sch{\"a}rdinger$^{*, 2}$, Murat Arcak$^{1}$, and Matthias Althoff$^{2}$
\thanks{$^{*}$ Equal contribution.}
 \thanks{$^{1}$ Hanna Krasowski and Murat Arcak are with the University of California, Berkeley, USA
         {\tt\small \{krasowski, arcak\}@berkeley.edu}}%
\thanks{$^{2}$ Stefan  Sch{\"a}rdinger and Matthias Altoff are with the School of Computation and Information Technology, Technical University of Munich, Munich, Germany
         {\tt\small \{stefan.schaerdinger, althoff\}@tum.de}}%
}

\markboth{}%
{Krasowski \MakeLowercase{\textit{et al.}}: Intelligent Sailing Model for Open Sea Navigation}

\IEEEpubid{}

\maketitle

\begin{abstract}

Autonomous vessels potentially enhance safety and reliability of seaborne trade. To facilitate the development of autonomous vessels, simulations are required to model realistic interactions with other vessels. 
However, modeling realistic interactive maritime traffic is challenging due to the unstructured environment, coarsely specified traffic rules, and largely varying vessel types. 
Currently, there is no standard for simulating interactive maritime environments in order to rigorously benchmark autonomous vessel algorithms. 
In this paper, we introduce the first intelligent sailing model (ISM), which simulates rule-compliant vessels for navigation on the open sea. An ISM vessel reacts to other traffic participants according to maritime traffic rules while at the same time solving a motion planning task characterized by waypoints.
In particular, the ISM monitors the applicable rules, generates rule-compliant waypoints accordingly, and utilizes a model predictive control for tracking the waypoints.
We evaluate the ISM in two environments: interactive traffic with only ISM vessels and mixed traffic where some vessel trajectories are from recorded real-world maritime traffic data or handcrafted for criticality. 
Our results show that simulations with many ISM vessels of different vessel types are rule-compliant and scalable. 
We tested 4,049 critical traffic scenarios. For interactive traffic with ISM vessels, no collisions occurred while goal-reaching rates of about 97 percent were achieved. 

\end{abstract}

\begin{IEEEkeywords}
Unmanned surface vessels, maritime autonomous surface ships, {COLREGS}, motion planning,  traffic simulation, collision avoidance, multi-agent systems.
\end{IEEEkeywords}

\section{Introduction}\label{sec:intro}
 
Autonomous navigation of surface vessels is a challenging application for autonomous systems \cite{Gu2021, Park2021,Qiao2023} as it features static and dynamic obstacles in an unstructured environment, varying environmental disturbances, and nonlinear vessel dynamics. Additionally, compared to mobile robot settings where collision avoidance behavior is commonly unconstrained, the collision avoidance interaction of surface vessels is explicitly defined by legally stipulated maritime traffic rules.

To efficiently develop and test autonomous systems, simulations that capture interactions of other agents with the autonomous system are essential. Examples include traffic simulations for autonomous driving \cite{Dosovitskiy.2017,lopez2018microscopic} and human-robot environments \cite{Inamura2021, Thumm2024}. 
For autonomous driving research, a variety of driver models \cite{Matcha2020} such as the Intelligent Driver Model (IDM) \cite{Treiber2000} have been developed to represent realistic and reactive traffic flow surrounding a system under test. 
The integration of the driver models in standard simulators such as CARLA \cite{Dosovitskiy.2017} or SUMO \cite{lopez2018microscopic} accelerates the development of autonomous road vehicles and is a significant asset for benchmarking new motion planning algorithms. 
For autonomous vessels, no comparable models exist even though the \ac{colregs} \cite{IMO.1972} specifies universally applicable traffic rules for collision-free navigation.

Many approaches for autonomous navigation consider the \ac{colregs}, e.g., in the cost function of an optimal control approach \cite{Johansen.2016, Kufoalor2019,  Tsolakis2022}, or in the reward of reinforcement learning \cite{Krasowski2024.safeRLautonomousVessels, Meyer2020}.
The effectiveness of single-agent approaches is usually evaluated on non-interactive traffic where the other traffic participants sail on a straight path \cite{Johansen.2016, Tsolakis2022} or are based on recorded traffic data \cite{Krasowski2024.safeRLautonomousVessels, Meyer2020}. Alternatively, a multi-agent setting is considered where commonly all vessels are controlled by the same algorithm with limited interpretability due to neural network controllers \cite{Cho2022, Wei2022} or cost functions implicitly considering traffic rules \cite{Grgicevic2024}. Due to expensive real-world testing, the algorithms are only rarely evaluated on hardware \cite{Kufoalor2019}.
To accelerate the development of autonomous surface vessels and allow for more robust testing on hardware, interactive and realistic simulation of maritime vessels is necessary.
\IEEEpubidadjcol

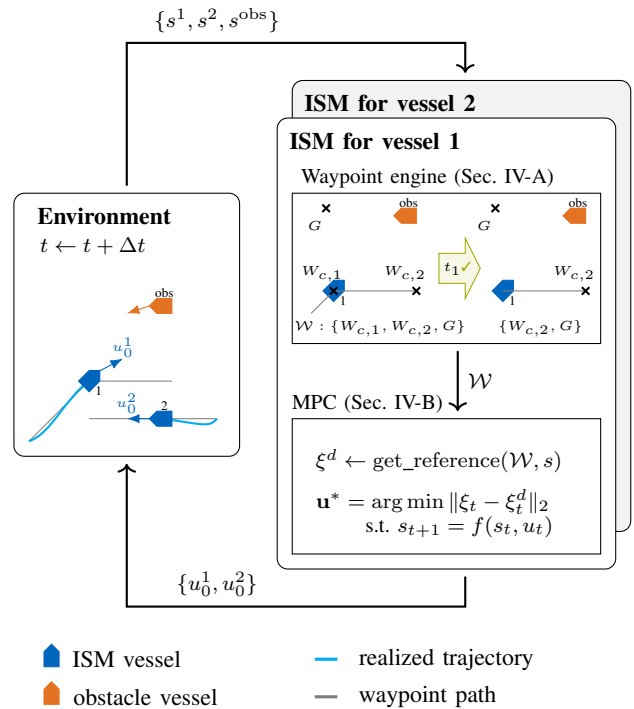
\begin{figure}
	\centering
	\def\vessel#1#2#3#4#5#6#7{
	\begin{scope}[shift={#1}, rotate=#2, scale=#3]
		\draw [fill=#4,draw=#5, opacity=#6](-1,1) -- (0,2) -- (1,1) -- (1,-1) -- (-1,-1) -- cycle;
		\node at (1.8,0) [text=black]{\tiny #7};
	\end{scope}
}

\def\waypoint#1#2#3#4{
	\begin{scope}[shift={#1}, scale=#3]
		\draw [thick](-0.1,-0.1) -- (0.1,0.1);
		\draw [thick](-0.1,0.1) -- (0.1,-0.1);
		\node at ($0.46*({cos(#2)},{sin(#2)}) - (0.3,0)$) {\tiny #4};	
	\end{scope}
}

\def\timemarker#1#2#3#4#5{
	\begin{scope}[shift={#1}, rotate=#2, scale=#3]
		\draw [thick](-0.2,0) -- (0.2,0);
		\node at ($0.3*({cos(#5)},{sin(#5)})$) {#4};
	\end{scope}
}

\def\timemarkerobs#1#2#3#4#5{
	\begin{scope}[shift={#1}, rotate=#2, scale=#3]
		\draw [thick, draw=TUMBlue](-0.1,0) -- (0.1,0);
		\node at ($0.3*({cos(#5)},{sin(#5)})$) [text=white] {#4};
	\end{scope}
}

\def\parallel#1{
	\begin{scope}[shift={#1}]
		\draw [thick](-0.1,-0.05) -- (0.1,0.05);
		\draw [thick](-0.1,0.05) -- (0.1,0.15);
	\end{scope}
}

\begin{tikzpicture}[node distance=4cm, auto]  
	\tikzset{
		mynode/.style={rectangle,rounded corners,draw=black,fill=TUMBlue!80,very thick, inner sep=1em, minimum size=3em, text centered, minimum width=10cm, text=white, minimum height=1.5cm},
		myarrow/.style={
			->,
			thick,
			shorten <=2pt,
			shorten >=2pt,},
		mylabel/.style={text width=7em, text centered} 
	} 
	
	
	\begin{scope}[yshift=2.5cm]
		\draw[rounded corners]  (0,0) rectangle (3,3.5);
		\node at (1.2,3.2) {\small \textbf{Environment}};
        \node at (1.05,2.8) {\footnotesize $t \leftarrow t + \Delta t$};
		\vessel{(2.0,2)}{90}{0.1}{TUMOrange}{TUMOrange}{1}{obs}
		\draw[gray] (0.2,0.2) -- (1,1) -- (2.1,1);
		\draw[thick, cyan] (0.2,0.2) .. controls (0.5,0.3) and (0.7,0.8).. (1,1);
		\draw[gray] (2.7,0.5) -- (1,0.5);
		\draw[thick, cyan] (2.7,0.5) .. controls (2.6,0.4) .. (2,0.5);
		\vessel{(2,0.5)}{90}{0.1}{TUMBlue}{TUMBlue}{1}{2}
		\vessel{(1,1)}{315}{0.1}{TUMBlue}{TUMBlue}{1}{1}
		\draw[TUMBlue, -{Latex}] (1.13,1.13) -- (1.45,1.3) node [above, yshift=-0.1cm] {\tiny $u_0^1$};
		\draw[TUMBlue, -{Latex}] (2,0.5) -- (1.5,0.5) node [above, yshift=-0.05cm] {\tiny $u_0^2$};
		\draw[TUMOrange, -{Latex}] (1.8,2) -- (1.5,1.9);
		
	\end{scope}
	
	\begin{scope}[yshift=1cm, xshift=4cm]
		\draw [fill=gray!10,rounded corners] (-0.3,0.5) rectangle (4.2,6.5);
		\node at (1.0,6.2) {\small \textbf{ISM for vessel 2}};
		\draw  [fill=white,rounded corners](-0.5,0) rectangle (4,6);
		\node at (0.8,5.7) {\small \textbf{ISM for vessel 1}};
		
		\node at (1.5,5.2) {\footnotesize Waypoint engine (Sec. \ref{subsec:waypoitnadaption})};
		\draw  (-0.3,3) rectangle (3.8,5.0);
		\begin{scope}[yshift=2.7cm, xshift=-0.75cm]
			
			\vessel{(2.0,2)}{90}{0.1}{TUMOrange}{TUMOrange}{1}{obs}
			\vessel{(1,1)}{315}{0.1}{TUMBlue}{TUMBlue}{1}{1}
			\draw[gray] (0.7,0.7) -- (1,1) -- (2.1,1);
			\waypoint{(1,1)}{90}{0.5}{$W_{c,1}$}
			\waypoint{(2.1,1)}{90}{0.5}{$W_{c,2}$}
			\waypoint{(0.9,2.1)}{270}{0.5}{$G$}
			\node at (1.62,0.5) {\tiny $\mathcal{W} : \{W_{c,1}, W_{c,2}, G\}$};
		\end{scope}
		\draw [fill=TUMGreen!10, draw=TUMGreen] (1.65,3.8) -- (2,3.8) -- (2,3.6) -- (2.2,4) -- (2,4.4)-- (2,4.2) -- (1.65,4.2) -- cycle;
		\node at (1.9,4) {\tiny $t_1$\textcolor{TUMGreen}{\ding{51}}};
		\begin{scope}[yshift=2.7cm, xshift=1.5cm]
			\vessel{(2.0,2)}{90}{0.1}{TUMOrange}{TUMOrange}{1}{obs}
			\vessel{(1,1)}{315}{0.1}{TUMBlue}{TUMBlue}{1}{1}
			\draw[gray] (1,1) -- (2.1,1);
			\waypoint{(2.1,1)}{90}{0.5}{$W_{c,2}$}
			\waypoint{(0.9,2.1)}{270}{0.5}{$G$}
			\node at (1.5,0.5) {\tiny $\{W_{c,2}, G\}$};
		\end{scope}
		\draw[thick, ->] (1.9,2.9) -- (1.9,2.1);
		\node at (2.2,2.5) {\footnotesize $\mathcal{W}$};

		\node at (0.7,2.2) {\footnotesize MPC (Sec. \ref{sc:opimitationmpcformulation})};
		\draw   (-0.3,0.2) rectangle (3.8,2.0);
		\node at (1.7,1) {\footnotesize \makecell[l]{$\mathbf{\xi}^d \leftarrow \mathrm{get\_reference}(\mathcal{W},s)$ \vspace{0.2cm} \\  $\mathbf{u}^* = \arg\min \|\xi_t - \xi_t^d \|_2 $ \\ $\qquad$ s.t. $s_{t+1} = f(s_t, u_t)$}};
		
	\end{scope}
	
	\draw[thick, ->] (6,0.9) -- (6,0.5) -- (1.5,0.5) -- (1.5,2.4);
	\node at (2.7,0.8) {\footnotesize $\{u_0^1, u_0^2\}$};
	\draw[thick, ->] (1.5,6.1) -- (1.5,8) -- (6,8)-- (6,7.6);
	\node at (2.7,8.3) {\footnotesize $\{s^1, s^2, s^\mathrm{obs}\}$};
	
	\begin{scope}[yshift=-0.7cm,xshift=0.5cm]
		\vessel{(0.0,0)}{0}{0.1}{TUMOrange}{TUMOrange}{1}{}
		\node at (1.25,0) {\small obstacle vessel};
		\vessel{(0,0.5)}{0}{0.1}{TUMBlue}{TUMBlue}{1}{}
		\node at (1,0.5) {\small ISM vessel};
		\draw[very thick, gray] (3.5,0.0) -- (3.8,0.0);
		\node at (5,0.0) {\small waypoint path};
		\draw[very thick, cyan] (3.5,0.5) -- (3.8,0.5);
		\node at (5.25,0.5) {\small realized trajectory};
	\end{scope}
	
\end{tikzpicture} 
	\caption{\ac{ism} integration with traffic environment. The depicted example consists of two \ac{ism} vessels and one obstacle vessel that is not reacting to the other vessels. Based on the states $s_i$, the \waypointadaption{} generates the set of waypoints $\mathcal{W}$ that characterize the motion planning task and potentially collision avoidance maneuver. In this example, the \ac{ism} vessel 1 is in a crossing situation at time step $t_1$ for which the waypoint $W_{c,1}$ is reached (see Fig.~\ref{fig:waypointadaption}) and the goal waypoint is $G$. For the 
    \ac{mpc}, the path characterized by the waypoint set $\mathcal{W}$ is transformed to a desired trajectory $\mathbf{\xi}^d$. Then, the optimal control input $\mathbf{u}^*$ is computed to track the reference given the vessel dynamics $f$.}
	\label{fig:headfigure}
\end{figure}

In this paper, we propose the first algorithm for modeling interpretable and interactive vessels on the open sea: the \ac{ism}. The \ac{ism} solves a motion planning problem while reacting to traffic participants according to maritime traffic rules (see Fig.~\ref{fig:headfigure}). We base the reactivity on formalized traffic rule predicates \cite{Krasowski2021.MarineTrafficRules} and introduce a \waypointadaption{} to generate the collision avoidance behavior specified in the \ac{colregs} and specify the motion planning task in the unstructured environment. The waypoints characterize the desired path that is tracked using a model predictive controller. This control architecture facilitates the application of the \ac{ism} for different vessel types. The \ac{ism} can be employed in multi-agent settings with an arbitrary number of both \ac{ism} vessels and vessels that are controlled by other motion planners. 
Our main contributions are:
\begin{itemize}
    \item We introduce the first interpretable parametrized sailing model for power-driven vessels that realistically and efficiently simulates maritime traffic on the open sea;
    \item Specifically, we propose a two-layer architecture with interpretable \waypointadaption{}, generating waypoints according to maritime traffic rules, and a lower-level \ac{mpc} that allows for independently adapting the vessel type and dynamics;
    \item We perform an extensive evaluation of the \ac{ism} with two vessel types, real-world traffic data, varying traffic density and criticality.
\end{itemize}

The remainder of this paper is structured as follows. In Section~\ref{sec:related work}, we present existing work on traffic models for vehicles and describe common setups for maritime traffic in maritime motion planning. Section~\ref{sec:preliminaries} introduces preliminary concepts and the problem statement. We detail the \ac{ism} in Section~\ref{sec:ISM} and evaluate its capabilities in Section~\ref{sec:experiments}. Last, we discuss our approach and conclude in Section~\ref{sec:discussion} and \ref{sec:conclusion}. 

\section{Related Work}\label{sec:related work}

\textbf{Traffic simulation. }
Developing safe and efficient autonomous vehicles is challenging since an infinite amount of traffic situations can occur on the road \cite{Hesham2021}, on the water \cite{Gu2021}, and in the air \cite{Hamissi2023}. Thus, computer simulations of traffic are important to accelerate and facilitate the development. These simulations are commonly based on statistical models derived from recorded real-world traffic or dynamical models for traffic participants. A major advantage of using dynamical models is that this allows for defining interactions between traffic participants and the autonomous vehicle. For road traffic, car-following models \cite{Matcha2020} such as the IDM \cite{Treiber2000} are commonly used to simulate traffic. The main idea of car-following models is that they react to preceding vehicles by maintaining a safe distance while aiming to drive at a desired velocity to ensure smooth traffic flow. For airborne traffic, encounter models are used to simulate traffic and evaluate collision avoidance methods \cite{Kochenderfer2010}. For maritime traffic, there are no standard models for simulating traffic. Instead, traffic participants based on recorded \cite{Krasowski2024.safeRLautonomousVessels,Meyer2020} or handcrafted trajectories \cite{Johansen.2016, Tsolakis2022} are commonly used. This work develops a parameterized vessel model, the \ac{ism}, that reacts to its surrounding traffic. Since the reactive behavior of the \ac{ism} is interpretably parametrized for different vessel types and follows maritime traffic rules, the \ac{ism} can be viewed as a maritime version of the IDM for road traffic. The most important difference between \ac{ism} and IDM arises from the unstructured maritime environment, where the \ac{ism} \waypointadaption{} determines reference paths while driver models like the IDM can leverage the road network to obtain a reference path.

\textbf{Unmanned surface vessels. }
Research on autonomous vessels, also called unmanned surface vessel (USV) or maritime autonomous surface ship (MASS), involves a diversity of task such as autonomous harbor operations \cite{Lexau2023}, inland waterway navigation \cite{YU2023-TSS, Vatle2024}, and long-distance routing \cite{Kim2023-routing, Kim2021-routing}, a diversity of vessel types such as small passenger ferries \cite{Wang2020roboat, Wang2019roboat} and large container ships \cite{felski2020ocean}, and a diversity of methods for perception \cite{Lin2022-perception, Pereira2021-perception, Yeong2021-perception}, planning \cite{Chiang2018, Enevoldsen2021, Lazarowska2016,  Stankiewicz2021, Zhang2022} and control \cite{Kufoalor2019, Johansen.2016, Tsolakis2022, Heiberg2022}. 
For the local planning and control, the surrounding vessels are an important factor for safe and efficient autonomous decision-making. This is particularly the case in dense traffic along shipping routes and for open sea encounters. In traffic situations with no static obstacles or waterway markings, only maritime traffic rules are providing structure to make decisions. Further, when evaluating autonomous agents, the interactions with the surrounding traffic play a critical role in performance and safety evaluation, making an easily sharable and reproducible traffic configuration desirable. 
Thus, we focus on these open sea navigation tasks for developing a reactive sailing model.

\textbf{COLREGS informed autonomous navigation. }
For navigation on the open sea, traffic rules specify reactive behavior. Similar to road traffic, traffic rules for maritime traffic depend on the jurisdiction. However, the most important collision avoidance rules are specified in the \ac{colregs} \cite{IMO.1972}. 
Recent research on \ac{colregs}-informed navigation of autonomous vessels can be categorized in deep reinforcement learning-based approaches \cite{Cho2022,Chun2021, Heiberg2022, Krasowski2024.safeRLautonomousVessels, Meyer2020, Wei2022, Xu2022} classical planning \cite{Chiang2018, Enevoldsen2021,HAN2024-planning,Stankiewicz2021, tengesdal2025comparative-planning, Zhang2022}, and control approaches \cite{ Enevoldsen2021, Grgicevic2024,Johansen.2016,Kufoalor2019, Li2025-APF, Lyu_Yin_2019-APF}. Due to their model-free nature, deep reinforcement learning-based approaches do not achieve the interpretability and adaptability required for an interactive vessel model. 
In contrast, the behaviors generated with classical planning and control approaches are more interpretable and predictable. For planning methods, rapidly-exploring random trees and A* search are common, and \ac{colregs} are integrated in the evaluation heuristic \cite{Stankiewicz2021, tengesdal2025comparative-planning, Zhang2022} or through virtual obstacles \cite{Chiang2018, Enevoldsen2021, HAN2024-planning}. For classical control approaches, \ac{mpc} \cite{Johansen.2016,Kufoalor2019, Tsolakis2022} and artificial potential fields \cite{Li2025-APF, Lyu_Yin_2019-APF} are dominating. To consider \ac{colregs}, existing \ac{mpc} approaches construct \ac{colregs}-informed cost functions and artificial potential field methods generate forces that pull vessels into \ac{colregs} compliant behavior. 
However, the heuristics or cost function do not explicitly specify the path of the vessel, but rather multiple components that need to be considered for computing a control input. Further, path planning methods with virtual obstacles do not explicitly specify a behavior and often plan for the full time horizon, which makes it difficult to predict trajectories and possible interactions along the path. To the best of our knowledge, our \ac{ism} is the first approach that generates interpretable and vessel-specific parameterized paths that comply with the \ac{colregs}, while generating the control signal in a separate step, which allows for different vessel types and dynamics fidelity.

\textbf{Monitoring with formal methods. }
Our \waypointadaption{} is related to runtime enforcement methods, where a specification is monitored and the control or system behavior is adapted depending on its satisfaction \cite{Konighofer2022, sanchez2019survey}. 
For safety requirements of systems operating in a continuous space, safeguards such as \ac{mpc}-based safety filters \cite{wabersich2023data} can be used to obtain probabilistic or hard safety guarantees. To achieve hard guarantees, the dynamics of the controlled system and the environment are modeled explicitly \cite{wabersich2021predictive} or implicitly, e.g., by control barrier functions \cite{taylor2020learning}. For autonomous vessels, Krasowski et al.~\cite{Krasowski2024.safeRLautonomousVessels} propose a safeguard that uses set-based reachability analysis to identify unsafe and safe actions and constrain an RL agent so that only safe actions are executed. In contrast, our \waypointadaption{} identifies if a maritime traffic rule is active and provides the necessary \adaptivewaypoints{} for a path that most likely complies with the traffic rules. This approach does not provide guarantees but is less computationally expensive and requires less modeling effort than monitoring with guarantees, since it is not necessary to model the dynamics of the environment. 

\section{Preliminaries} \label{sec:preliminaries}
We briefly introduce our notation and vessel model, the collision avoidance rules relevant for the \ac{ism}, and the regarded motion planning problem to state the problem investigated in this study.

\paragraph{Notation and vessel dynamics}
The state of a vessel is characterized by $\state = [\xposition, \yposition, \orientation, \nvelocity]$ with $\xposition \in \mathbb{R}$ and $\yposition\in \mathbb{R}$ being Cartesian coordinates of the current position $\positions$ of the vessel, the orientation $\orientation$ and the velocity $\nvelocity \in \mathbb{R}$ in the direction of $\orientation \in [-\pi, \pi]$.  The velocity vector $\mathbf{v}$ is defined by the magnitude $\nvelocity$ and the orientation $\orientation$. We use subscripts for the state and state components to specify a vessel. The dynamics for the vessels follow the yaw-constrained model \cite[Eq. (3)]{Krasowski2022.CommonOcean}:
\begin{align} \label{eq:yawconstrainedmodel}
f(\state, \control) =
    \begin{pmatrix}
\dot{\xposition}\\
\dot{\yposition}\\
\dot{\orientation}\\
\dot{\nvelocity}
\end{pmatrix} \ =\ \begin{pmatrix}
\cos( \orientation ) \cdotp \nvelocity\\
\sin( \orientation ) \cdotp \nvelocity\\
\turningrate \\
\nacceleration
\end{pmatrix},
\end{align}
where $\nacceleration \in \mathbb{R}$ is the acceleration in the direction of $\orientation$ and $\turningrate \in \mathbb{R}$ is the yaw of the vessel. The control input $\control$ is $[\nacceleration , \turningrate]$. Additionally, let us define the operator $\mathtt{proj}_{\triangle}$ projects the state to the dimension specified by $\triangle$ and the function $\mathtt{vec2rad} (\positions_{1}, \positions_{2}): \mathbb{R}^2 \times \mathbb{R}^2 \rightarrow [-\pi, \pi]$ computes the orientation in the Cartesian coordinate system of the vector $\positions_{1} - \positions_{2}$.
The function $\mathtt{rad2vec}: [-\pi, \pi] \rightarrow \mathbb{R}^2$ computes the normal vector of the angle $\alpha$:
\begin{equation}
    \mathtt{rad2vec}(\alpha) = [\cos(\alpha), \sin(\alpha)].
\end{equation}
The orthonormal projection of a vector $x \in \mathbb{R}^2$ onto a line defined by the non-zero vector $y \in \mathbb{R}^2$ is defined by
\begin{equation}
    \mathtt{orthop}(x, y) = \frac{x \cdot y}{y \cdot y} \, y.
\end{equation}
Further, let $g$ be a line in $\mathbb{R}^2$ defined by a point on the line $g_\positions$ and a vector along the line $g_\orientation$, in brief, $\langle g_\positions, g_\orientation \rangle$. Now, let the function $\mathtt{intersect}(g_1, g_2)$ calculate the point of intersection of the lines and return $\mathbf{0}$ if the lines are parallel. 
Additionally, we denote the distance traveled between two time points $t_0$ and $t_1$ by $d_{[t_0, t_1]}$. In a discrete-time setting, this is simply the sum of the Euclidean distances between all time steps in the time interval $[t_0, t_1]$.
Lastly, let us introduce the predicate $\mathrm{in\_h}(\positions, \beta^h, b)$ which evaluates to true if:
\begin{align}
	\mathrm{in\_h}(\positions, \beta^h, b) \iff \mathtt{rad2vec}(\beta^h) \, \positions - b \leq 0, \label{eq:halfspace}
\end{align}
where $\positions$ is a position in the Cartesian coordinate system, $\beta^h$ is the orientation perpendicular to the halfspace, and $b$ is an optional parameter to check if a position has distance $b$ to the halfspace boundary, i.e., it is set to zero if any position in the halfspace should be considered as within. 

\paragraph{Collision avoidance traffic rules}
For collision avoidance, the \ac{colregs} \cite{IMO.1972} differentiate between different vessel classes, such as sailing vessel, fishing vessel, and power-driven vessel. Since power-driven are the most common vessel class, we consider the traffic rules for collision avoidance of these vessels.
Specifically, there are three encounter situations specified in the \ac{colregs} for which appropriate collision avoidance maneuvers have to be performed: head-on, overtaking, and crossing (see Fig.~\ref{fig:waypointadaption}). In an encounter situation, a vessel can be a give-way vessel or a stand-on vessel. A give-way vessel has the obligation to evade while noticeably turning so that the other traffic participants can detect the evasion maneuver. A stand-on vessel has to keep its course and speed until the encounter situation is resolved. To evaluate compliance, the encounter traffic rules have been formalized with temporal logic \cite{ Krasowski2024.safeRLautonomousVessels,Krasowski2021.MarineTrafficRules,Torben2023}. In this work, we employ the formalized predicates from \cite{Krasowski2024.safeRLautonomousVessels} to detect encounter situations. Our \ac{ism} is designed to consider the encounter traffic rules $\mathcal{R} = \{ R_3, R_4, R_5, R_6\}$ \cite[Table I]{Krasowski2024.safeRLautonomousVessels}, where $R_3, R_4$, and $R_5$ are the formalizations for the give-way vessels in crossing, head-on, and overtaking encounters, respectively. The rule $R_6$ formalizes stand-on vessels. We refer the interested reader to \cite{Krasowski2024.safeRLautonomousVessels} for details on the formalized rules.

Nevertheless, let us briefly introduce the predicates from \cite{Krasowski2024.safeRLautonomousVessels} that relevant for the \ac{ism}. The traffic rules are collision avoidance rules and thus only need to be applied when a collision is possible. To detect a collision between two vessels with current states $\state_l$ and $\state_m$, we use the predicate $\mathrm{collision\_possible}$:
\begin{align}
	&\mathrm{collision\_possible}(\state_l, \state_m, t_\mathrm{horizon}) \iff \notag \\ & \quad \mathcal{V}_l \subseteq CC'(\state_l, \state_m) \, \land \notag \\
	& \quad \| \mathbf{v}_l - \mathbf{v}_m\|_2 \geq \|\mathtt{proj}_{\positions} (\state_l) - \mathtt{proj}_{\positions} (\state_m)\|_2 / t_\mathrm{horizon}, \label{eq:coll-possible}
\end{align}
where $t_\mathrm{horizon}$ is the prediction horizon, and $\mathcal{V}_l$ is the set of velocities to be checked for intersection with the velocity obstacle $CC'$ \cite{Fiorini1998}. Intuitively, a collision between vessel $l$ and $m$ is possible given the velocities remain constant if the velocity obstacle intersects with the set of velocities of vessel $l$ and the time to collision is less than $t_\mathrm{horizon}$.

Next, the three predicates for detecting give-way vessels are defined. First, a head-on give-way vessel is detected if:
\begin{align}
	&\mathrm{head\_on}(\state_l, \state_m, t_\mathrm{horizon},\Delta_{\mathrm{head\text{-}on}}) \iff \notag \\
	& \quad \mathrm{collision\_possible}(\state_l, \state_m, t_\mathrm{horizon}) \, \land \notag \\  & \quad \mathrm{in\_front\_sector} (\state_l, \state_m) \, \land \notag \\ 
	& \quad \lnot \mathrm{orientation\_delta}(\state_l, \state_m, \Delta_{\mathrm{head\text{-}on}}, \pi), \label{eq:head-on}
\end{align}
where $\Delta_{\mathrm{head\text{-}on}}$ is the parameter specifying the head-on sector, i.e., usually 5 or 10 degree, the predicate $\mathrm{in\_front\_sector}$ and $ \lnot \mathrm{orientation\_delta}$ evaluate if vessel $m$ is in a relative position and orientations with respect to the vessel $l$ so that the head-on rule applies.
Second, a crossing give-way vessel is identified if:
\begin{align}
	&\mathrm{crossing}(\state_l, \state_m, t_\mathrm{horizon},\Delta_{\mathrm{head\text{-}on}}) \iff \notag \\
	& \quad \mathrm{collision\_possible}(\state_l, \state_m, t_\mathrm{horizon}) \land \notag \\ & \quad \mathrm{in\_right\_sector} (\state_l, \state_m) \land \notag \\
	& \quad \mathrm{orientation\_towards\_left} (\state_l, \state_m, \Delta_{\mathrm{head\text{-}on}}), \label{eq:crossing}
\end{align}
where the predicate $\mathrm{in\_right\_sector}$ and $\mathrm{orientation\_towards\_left}$ check if vessel $m$ is at a relative position and orientation with respect to the vessel $l$ so that the crossing rule applies. Third, for an overtaking give-way vessel, again a collision possibility has to exist and the relative position and orientation has to match the \ac{colregs}, i.e., $\mathrm{in\_behind\_sector}$ and $\lnot \mathrm{orientation\_delta}$ are fulfilled, while additionally the vessel $l$ has to sail faster than the other vessel $m$:
\begin{align}
	&\mathrm{overtake}(\state_l, \state_m, t_\mathrm{horizon}) \iff \notag \\
	& \quad \mathrm{collision\_possible}(\state_l, \state_m, t_\mathrm{horizon}) \, \land \notag \\ & \quad \mathrm{in\_behind\_sector} (\state_m, \state_l) \, \land \notag \\  
	&\quad \lnot \mathrm{orientation\_delta}(\state_l, \state_m, \SI{67.5}{\degree}, 0) \, \land  \notag \\
     &\quad \mathrm{sails\_faster}(\state_l, \state_m), \label{eq:overtake}
\end{align}
where the predicate $\mathrm{sails\_faster}$ detects if the velocity of vessel $l$ is significantly above the velocity of vessel $m$. 
Lastly, stand-on vessels have to keep their course and speed and are present in crossing and overtaking encounters (see Fig.~\ref{fig:waypointadaption}). The predicate $ \mathrm{keep}$ for detecting a stand-on vessel $l$ with respect to another vessel $m$ is:
\begin{align}
	&\mathrm{keep}(\state_l, \state_m, t_\mathrm{horizon},\Delta_{\mathrm{head\text{-}on}}) \iff \notag \\
	& \quad \big(\mathrm{collision\_possible}(\state_l, \state_m, t_\mathrm{horizon}^\mathrm{check}) \land \notag \\ & \quad \mathrm{in\_left\_sector} (\state_l, \state_m) \land \notag \\ & \quad \mathrm{orientation\_towards\_right} (\state_l, \state_m, \Delta_{\mathrm{head\text{-}on}}) \big) \lor \notag \\ & \quad \mathrm{overtake}(\state_m, \state_l, t_\mathrm{horizon}^\mathrm{check}), \label{eq:keep}
\end{align}
where the first part of the disjunction detects the stand-on vessel in crossing encounters and the second part detects an overtaking encounter. 
For a detailed introduction of the rules and predicates, we refer the interested reader to \cite{Krasowski2024.safeRLautonomousVessels, Krasowski2021.MarineTrafficRules}.

\paragraph{Motion planning on the open sea}
For road traffic, the behavior of traffic participants is structured by the road network. In contrast, vessel navigation on the open sea is unstructured.
Consequently, motion planning tasks for vessels can be defined as $(\state_0, \mathcal{G})$ where $\state_0$ is the initial state, and $\mathcal{G}$ is an ordered list of waypoints that includes the goal state, also called goal waypoint, as last waypoint. The intermediate waypoints characterize necessary course changes.

\paragraph{Modeling goals}
For modeling realistic maritime traffic, a vessel has to react to traffic participants according to maritime traffic rules while solving a motion planning task. 
To reduce complexity, we assume that all traffic participants are power-driven vessels, and the traffic situation  
is on the open sea so that the interactive traffic rules are reduced to the rule set $\mathcal{R}$. Note that the maneuver-of-the-last-minute rule ($R_1$ in \cite{Krasowski2024.safeRLautonomousVessels}) and the safe speed rule ($R_2$ in \cite{Krasowski2024.safeRLautonomousVessels}) are not implemented since the maneuver-of-the-last-minute rule is not necessary if all vessels follow the collision-avoidance rules $\mathcal{R}$ and the safe speed rule can be trivially ensured by adjusting the maximum velocity. 

To obtain realistic traffic on the open sea, the problem is to identify a model $\mathcal{M}$ with parameters $\Gamma$ that produces a trajectory $\xi$, which is compliant with the rule set $\mathcal{R}$. Additionally, the model should minimize the deviation from the desired trajectory $\xi^d$ characterized by a motion planning task $(\state_0, \mathcal{G})$:
\begin{align}
    \min \sum_{i=0}^{T} \| \state_i - \state^d_i \|_2
\end{align}
where $T$ is the maximum time step of the trajectory, $\state_i$ and $\state^d_i$ is that state at time step $i$ of the trajectory $\xi$ and $\xi^d$, respectively. 
The rule-compliant interactions have a higher priority than solving the motion planning task $(\state_0, \mathcal{G})$ accurately. The model $\mathcal{M}$ should have the following features: 
\begin{itemize}
    \item Interpretablity to ease validation and design of \ac{colregs} behavior by experts and authorities;
    \item Seamless adjustability so that it can be used for different traffic modes of different locations or times of the day;
    \item Modularity so that different vessel types can be used;
    \item Computational efficiency so that it is scalable to high traffic density and a viable solution for extensive testing of autonomous vessel algorithms.
    \item Realistic simulation frequency of 0.5 - 2 Hz should be computable faster than or at real time.
\end{itemize}

\section{Intelligent Sailing Model} \label{sec:ISM}

The \ac{ism} describes a closed-loop system of a vessel with a two-step control approach: a \waypointadaption{} for traffic-reactive behavior introduced in Sec.~\ref{subsec:waypoitnadaption} and a model predictive controller to track the path characterized by the waypoints introduced in Sec.~\ref{sc:opimitationmpcformulation} (see Fig.~\ref{fig:headfigure}). At each time step, the \waypointadaption{} checks if a new encounter situation is present, the \adaptivewaypoints{} need to be adjusted to guide the vessel along a traffic rule-compliant collision avoidance path, or the encounter is resolved and the vessel can continue to solve the motion planning problem initially provided. To bridge the gap between the sparse waypoints and a time-discrete control signal for a specific vessel, the waypoints are transformed to a reference trajectory that is tracked with a model predictive controller, which considers the vessel dynamics as constraints. An alternative approach may be to design a velocity control along a spline, which has the waypoints as control points. However, the \ac{mpc} for waypoint tracking is better adaptable for different vessel dynamics and can be more easily extended to consider environmental disturbances.

\subsection{Rule-adaptive waypoint engine for collision avoidance} \label{subsec:waypoitnadaption}

The \ac{ism} vessel is initialized with a motion planning task $(\state_0, \mathcal{G}_i)$ where $\mathcal{G}_i$ is the initial waypoint list. During runtime, the \ac{ism} vessel monitors if other vessels in the traffic situation require it to react according to the maritime traffic rules. In particular, four cases can be observed: the \ac{ism} vessel can be the give-way vessel for head-on, overtaking, and crossing encounters or the stand-on vessel, which are detected with the predicates $\mathrm{head\_on}, \mathrm{crossing}, \mathrm{overtake}$ and $\mathrm{keep}$, respectively (see Eq.~\eqref{eq:head-on} -- \eqref{eq:keep}).
For give-way vessels, the encounter predicate needs to be detected for at least the reaction time $t_\mathrm{react}$ before a collision avoidance maneuver is started, i.e., the path adaptation is only performed if the encounter predicate is present for $t_\mathrm{react}$. Since the encounter predicates are mutually exclusive (see \cite[Lemma 1]{Krasowski2024.safeRLautonomousVessels}), it is ensured that two encounter situations cannot be detected at the same time. Given an encounter situation, the waypoint adaption is performed as described in the following paragraphs for the four different encounters. For the adaption, we define two types of waypoints: \emph{normal waypoints} and \emph{guiding waypoints}. A normal waypoint $W$ has to be approached by at least the distance $d_\mathrm{wp}$:
\begin{equation}\label{eq:waypointreached}
    \| \positions - W\|_2 \leq d_\mathrm{wp}.
\end{equation}
In contrast, a guiding waypoint is positioned sufficiently far from the vessel so that it cannot be reached within the prediction horizon of the model predictive controller and, thus, directs the vessel in a specific direction. A sufficient distance is specific to a vessel type and can be determined based on the maximum velocity. Note that guiding waypoints are a way to achieve desired orientation tracking without the need to change the cost function of the \ac{mpc} problem (see \eqref{eq:MPC}).

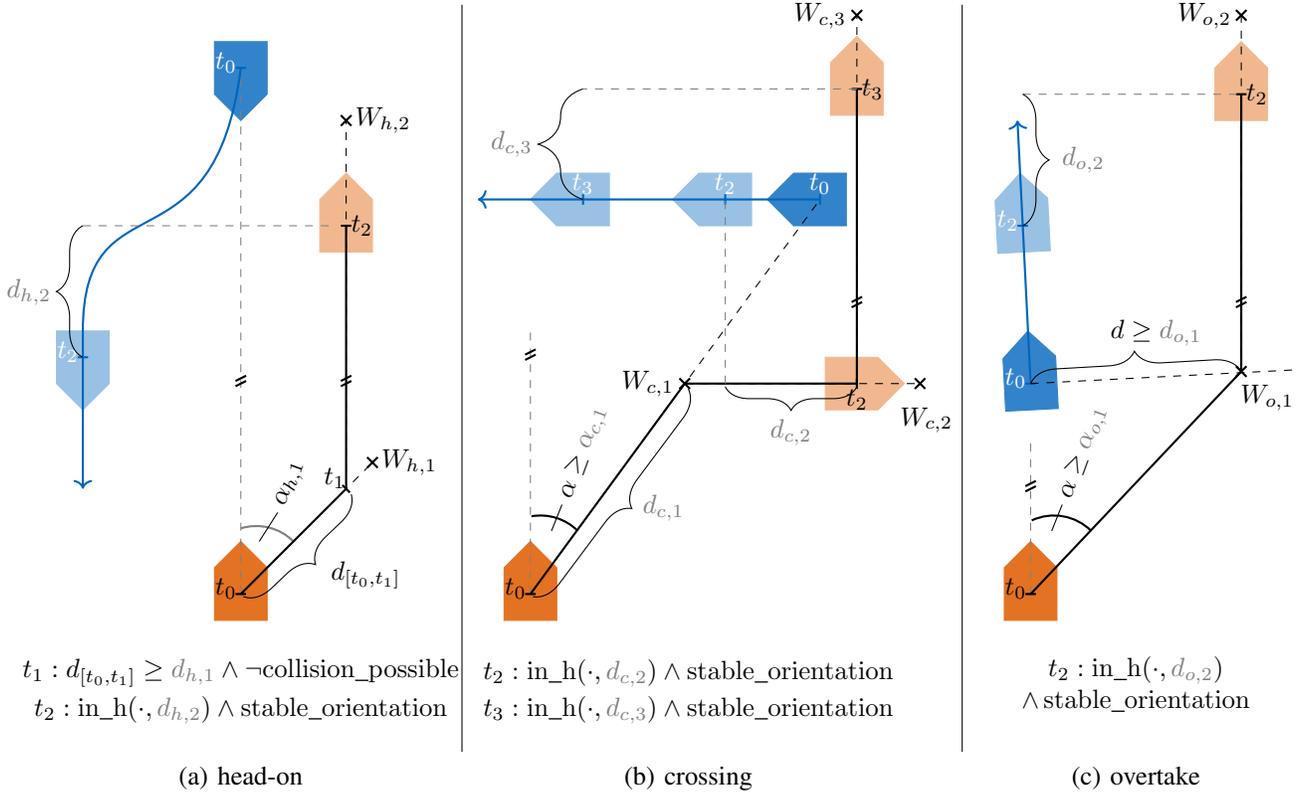
\begin{figure*}
	\vspace{0.2cm}
	\centering
	\def\vessel#1#2#3#4#5#6{
	\begin{scope}[shift={#1}, rotate=#2, scale=#3]
		\draw [fill=#4,draw=#5, opacity=#6](-1,1) -- (0,2) -- (1,1) -- (1,-1) -- (-1,-1) -- cycle;
	\end{scope}
}

\def\waypoint#1#2#3#4{
	\begin{scope}[shift={#1}, scale=#3]
		\draw [thick](-0.1,-0.1) -- (0.1,0.1);
		\draw [thick](-0.1,0.1) -- (0.1,-0.1);
		\node at ($0.7*({cos(#2)},{sin(#2)})$) {#4};	
	\end{scope}
}

\def\timemarker#1#2#3#4#5{
	\begin{scope}[shift={#1}, rotate=#2, scale=#3]
		\draw [thick](-0.1,0) -- (0.1,0);
		\node at ($0.3*({cos(#5)},{sin(#5)})$) {#4};
	\end{scope}
}

\def\timemarkerdiff#1#2#3#4#5#6{
	\begin{scope}[shift={#1}, rotate=#2, scale=#3]
		\draw [thick](-0.1,0) -- (0.1,0);
		\node at ($#6*({cos(#5)},{sin(#5)})$) {#4};
	\end{scope}
}

\def\timemarkerobs#1#2#3#4#5{
	\begin{scope}[shift={#1}, rotate=#2, scale=#3]
		\draw [thick, draw=TUMBlue](-0.1,0) -- (0.1,0);
		\node at ($0.3*({cos(#5)},{sin(#5)})$) [text=white] {#4};
	\end{scope}
}

\def\parallel#1{
	\begin{scope}[shift={#1}]
		\draw [thick](-0.1,-0.05) -- (0.1,0.05);
		\draw [thick](-0.1,0.05) -- (0.1,0.15);
	\end{scope}
}

\begin{tikzpicture}[node distance=4cm, auto, scale=0.7]  
	\tikzset{
		mynode/.style={rectangle,rounded corners,draw=black,fill=TUMBlue!80,very thick, inner sep=1em, minimum size=3em, text centered, minimum width=10cm, text=white, minimum height=1.5cm},
		myarrow/.style={
			->,
			thick,
			shorten <=2pt,
			shorten >=2pt,},
		mylabel/.style={text width=7em, text centered} 
	} 
	
	
	\begin{scope}
	
	\vessel{(0,0)}{0}{0.5}{TUMOrange}{TUMOrange}{1}
	\vessel{(0,10)}{180}{0.5}{TUMBlue}{TUMBlue}{0.8}
	\vessel{(-3,4.5)}{180}{0.5}{TUMBlue}{TUMBlue}{0.4}
	\vessel{(2,7)}{0}{0.5}{TUMOrange}{TUMOrange}{0.5}
	
	\draw[gray,thick] (1,1) arc[start angle=45, end angle=107, radius=1cm];
    \node at (0.95, 2.1) [rotate=70] {\textcolor{gray}{$\alpha_{h,1}$}};
	\draw (0.35,1.0) --(0.7,1.55);
	\draw[dashed, gray] (0,0) -- (0,10);
	\parallel{(0,4)}
	\timemarker{(0,0)}{0}{1}{$t_0$}{160}
	
	\draw[thick, TUMBlue, ->] (0,10) 
	.. controls (-0.5,6)  and (-3,8) .. (-3,5) -- (-3,2);
	\timemarkerobs{(-3,4.5)}{0}{1}{$t_2$}{160}
	\timemarkerobs{(0,10)}{0}{1}{$t_0$}{160}
	
	\draw[dashed] (0,0) -- (2.5,2.5);
	\draw[thick] (0,0) -- (2,2);
	\waypoint{(2.5,2.5)}{0}{1}{$W_{h,1}$}
	\timemarker{(2,2)}{-45}{1}{$t_1$}{180}
	\draw[decorate,decoration={brace,amplitude=10pt, mirror}] (0.05,-0.05) -- (2.1,1.9) node[midway,xshift=1.5cm, yshift=-0.7cm] {$d_{[t_0,t_1]}$};
	
	\draw[dashed] (2,2) -- (2,9);
	\draw[thick] (2,2) -- (2,7);
	\parallel{(2,4)}
	\waypoint{(2,9)}{0}{1}{$W_{h,2}$}
	
	\timemarker{(2,7)}{0}{1}{$t_2$}{0}
	\draw[decorate,decoration={brace,amplitude=10pt}] (-3,4.5) -- (-3,7) 
	node[midway,xshift=-0.3cm] {$\textcolor{gray}{d_{h,2}}$};
    
    \draw[dashed, gray] (-3,7) -- (1.8,7) ;
	
	\node at (0,-1.5) {$t_1: d_{[t_0,t_1]} \geq \textcolor{gray}{d_{h,1}} \land \lnot \mathrm{collision\_possible}$};
	\node at (0,-2.2) {$t_2: \mathrm{in\_h}(\cdot, \textcolor{gray}{d_{h,2}}) \land  \mathrm{stable\_orientation}$};

    \node at (0,-3.5) {(a) head-on};
	\end{scope}
	\draw (4.2,-3) -- (4.2,11.2);
	\begin{scope}[xshift=8.5cm]
		
		\vessel{(-3,0)}{0}{0.5}{TUMOrange}{TUMOrange}{1}
		\vessel{(2.5,7.5)}{90}{0.5}{TUMBlue}{TUMBlue}{0.8}
		
		\vessel{(0.7,7.5)}{90}{0.5}{TUMBlue}{TUMBlue}{0.4}
		\vessel{(3.1,4)}{270}{0.5}{TUMOrange}{TUMOrange}{0.5}
		\vessel{(-2.0,7.5)}{90}{0.5}{TUMBlue}{TUMBlue}{0.4}
		\vessel{(3.2,9.6)}{0}{0.5}{TUMOrange}{TUMOrange}{0.5}
		
		\draw[thick] (-2.1,1.2) arc[start angle=45, end angle=100, radius=1cm];
		\node at (-2.0, 2.7) [rotate=70] {$\alpha \geq \textcolor{gray}{\alpha_{c,1}}$};
		\draw (-2.6,1.2) --(-2.4,1.7);
		\draw[dashed, gray] (-3,0) -- (-3,5);
		\parallel{(-3,4.5)}
		\timemarker{(-3,0)}{0}{1}{$t_0$}{160}
		
		\draw[thick, TUMBlue, ->] (2.5,7.5) -- (-4,7.5);
		\timemarkerobs{(-2.0,7.5)}{90}{1}{$t_3$}{0}
		\timemarkerobs{(0.7,7.5)}{90}{1}{$t_2$}{0}
		\timemarkerobs{(2.5,7.5)}{90}{1}{$t_0$}{0}
		
		\draw[dashed, black] (-3,0) -- (2.5,7.5);
		\draw[thick] (-3,0) -- (-0.07,4);
		\waypoint{(-0.07,4)}{180}{1}{$W_{c,1}$}
		\draw[decorate,decoration={brace,amplitude=10pt, mirror}] (-2.95,-0.05) -- (0,3.9) node[midway,xshift=1.1cm, yshift=-0.5cm] {$\textcolor{gray}{d_{c,1}}$}; 
		
		\draw[dashed, black] (-0.07,4) -- (4.4,4);
		\draw[thick] (-0.07,4) -- (3.2,4);
		\parallel{(3.2,5.5)}
		\waypoint{(4.4,4)}{-80}{1}{$W_{c,2}$}
		
		\timemarker{(3.2,4)}{-90}{1}{$t_2$}{0}
        
  \draw[decorate,decoration={brace,amplitude=9pt, mirror}] (0.7,3.94) -- (3.2,3.94) 
		node[midway,yshift=-0.9cm] {$\textcolor{gray}{ d_{c,2}}$};
        \draw[dashed, gray] (0.7,4) -- (0.7,7.5) ;
		
		\draw[dashed, black] (3.2,4) -- (3.2,11);
		\draw[thick] (3.2,4) -- (3.2,9.6);
		\waypoint{(3.2,11)}{180}{1}{$W_{c,3}$}
		\timemarker{(3.2,9.6)}{0}{1}{$t_3$}{0}
		\draw[decorate,decoration={brace,amplitude=17pt}] (-2.0,7.5) -- (-2.0,9.6) 
		node[midway,xshift=-0.55cm] {$\textcolor{gray}{d_{c,3}}$};
        \draw[dashed, gray] (3.5,9.6) -- (-2.0,9.6);
		
		\node at (0,-1.5) {$t_2: \mathrm{in\_h}(\cdot,\textcolor{gray}{ d_{c,2}}) \land  \mathrm{stable\_orientation}$};
		\node at (0,-2.2) {$t_3: \mathrm{in\_h}(\cdot, \textcolor{gray}{d_{c,3}}) \land  \mathrm{stable\_orientation}$};

    \node at (0,-3.5) {(b) crossing};
	\end{scope}
	\draw (13.7,-3) -- (13.7,11.2);
	\begin{scope}[xshift=17cm]
		
		\vessel{(-2,0)}{0}{0.5}{TUMOrange}{TUMOrange}{1}
		\vessel{(-2,4)}{3}{0.5}{TUMBlue}{TUMBlue}{0.8}
		\vessel{(-2.15,7)}{3}{0.5}{TUMBlue}{TUMBlue}{0.4}
		\vessel{(2,9.5)}{0}{0.5}{TUMOrange}{TUMOrange}{0.5}
		
		\draw[thick] (-0.85,1.2) arc[start angle=45, end angle=115, radius=1cm];
		\node at (-1.0, 2.7) [rotate=70] {$\alpha \geq \textcolor{gray}{\alpha_{o,1}}$};
		\draw (-1.6,1.2) --(-1.4,1.7);
		\draw[dashed, gray] (-2,0) -- (-2,3);
		\parallel{(-2,2)}
		\timemarker{(-2,0)}{0}{1}{$t_0$}{160}
		
		\draw[thick, TUMBlue, ->] (-2,4) -- (-2.25,9);
		\timemarkerobs{(-2.15,7)}{0}{1}{$t_2$}{160}
		\timemarkerobs{(-2,4)}{0}{1}{$t_0$}{160}
		
		\draw[dashed] (-2,4) -- (3,4.26);
		\draw[thick] (-2,0) -- (2,4.23);
		\waypoint{(2,4.23)}{-45}{1}{$W_{o,1}$}
		\draw[decorate,decoration={brace,amplitude=10pt}] (-2,4) -- (2,4.23) node[midway, xshift=1cm, yshift=0.3cm] {$d \geq \textcolor{gray}{d_{o,1}}$};
		
		\draw[dashed] (2,4.23) -- (2,11);
		\draw[thick] (2,4.23) -- (2,9.5);
		\parallel{(2,5.5)}
		\waypoint{(2,11)}{180}{1}{$W_{o,2}$}
		\timemarker{(2,9.5)}{0}{1}{$t_2$}{0}
		
		\draw[dashed, gray] (-2.15,9.5) -- (2,9.5); 
		\draw[decorate,decoration={brace,amplitude=10pt}] (-2.15,9.5) -- (-2.15,7)
		node[midway,xshift=0.4cm] {$\textcolor{gray}{ d_{o,2}}$};
		\node at (0,-1.75) {\makecell[c]{$t_2: \mathrm{in\_h}(\cdot,\textcolor{gray}{ d_{o,2}})$ \\ $ \land  \, \mathrm{stable\_orientation}$}};
        
        \node at (0,-3.5) {(c) overtake};
	\end{scope}
\end{tikzpicture} 
	\caption{Waypoint adaption for collision avoidance maneuvers of an \ac{ism} vessel in head-on, crossing, and overtake situations as the give-way vessel. The obstacle vessel and its rule-compliant path are depicted in blue, and the \ac{ism} vessel is in orange, with the desired avoidance path as a solid black line. Tunable parameters are gray.}
	\label{fig:waypointadaption}
\end{figure*}

\paragraph{Stand-on}
If the \ac{ism} vessel is detected to be a stand-on vessel via the predicate $\mathrm{keep}$ (see Eq.~\ref{eq:keep}), it has to maintain its course and speed. Thus, we trivially adapt the path to a single guiding waypoint in the direction of the current heading of the vessel $\mathtt{proj}_{\orientation}(\state_k)$, where $\state_k$ is the state of the vessel when $\mathrm{keep}$ first evaluates to true. Once $\mathrm{keep}$ evaluates to false, the vessel resumes its original path as detailed in Subsection \ref{subsec:waypoitnadaption}.

\paragraph{Head-on}
In a head-on encounter, a vessel has to evade to the right, pass the other vessel, and then return to its original path. To achieve this, we interactively adjust the waypoints to be tracked in two phases (see Fig.~\ref{fig:waypointadaption}(a)).
First, the guiding waypoint $W_{h,1}$ is created on the line defined by the orientation $\mathtt{proj}_{\orientation} (\state_{\mathrm{ism},0}) - \alpha_\mathrm{h,1}$ and the current position of the vessel $\mathtt{proj}_{\positions} (\state_{\mathrm{ism},0})$ (where the second subscript $0$ indicates the time point $t_0$) so that the vessel turns by $\alpha_\mathrm{h,1}$ to the right. The vessel follows this direction until the time point $t_1$, when the condition $\lnot \mathrm{collision\_possible} \land d_{[0,t]} \geq d_{h, 1}$ is fulfilled (where the predicate $\mathrm{collision\_possible}$ is defined in Eq.~\eqref{eq:coll-possible}). At $t_1$, the guiding waypoint $W_{h,2}$ is calculated so that the vessel starts sailing parallel to the line connecting both vessels' positions at $t_0$ (i.e., the direction is $ \mathtt{proj}_{\positions} (\state_{\mathrm{obs},0}) - \mathtt{proj}_{\positions} (\state_{\mathrm{ism},0})$). The waypoint $W_{h,2}$ is tracked until the other vessel is at least $d_{ho, 2}$ behind the \ac{ism} vessel and its orientation is stable toward $W_{h,2}$. A stable orientation is present if the orientation difference to the desired orientation is smaller than $\alpha_\mathrm{so}$ for at least $t_\mathrm{so}$:
\begin{align}
    &\mathrm{stable\_orientation}(\xi, \positions_1, \positions_{2}) \iff \\ & \;\; |\mathtt{proj}_\orientation(\state_t) - \mathtt{vec2rad}(\positions_1, \positions_{2}) | \leq \alpha_\mathrm{so} \; \forall t \in [-t_\mathrm{so},0].\notag
\end{align}
For the head-on collision avoidance, the arguments $\positions_1$ and  $\positions_2$, are the position of the vessel at $t_1$ and the waypoint $W_{h,2}$. Once the condition for $t_{2}$ is fulfilled, the vessel returns to its original path.
The halfspace predicate \eqref{eq:halfspace} evaluates if the vessel is far enough behind using the following parameters:
\begin{align}
    \mathrm{in\_h}(\mathtt{proj}_{\positions}(\state_\mathrm{obs}), \mathtt{proj}_{\orientation}(\state_\mathrm{ism}), d_{h,2}).
\end{align}
Note that the halfspace is shifted from the origin of the Cartesian coordinate system so that the position of the \ac{ism} vessel $\mathtt{proj}_{\positions}(\state_\mathrm{ism})$ is on the halfspace boundary. Since we use this predicate in the other give-way encounters with the same first three arguments as well, let us abbreviate it with $\mathrm{in\_h}(\cdot, d_{h,2})$.

\paragraph{Crossing}
In a crossing encounter, the give-way vessel has to evade to the right by significantly changing its orientation and should pass behind the other vessel (see Fig.~\ref{fig:waypointadaption}(b)). Thus, the first normal waypoint $W_{c,1}$ is generated based on the position of the other vessel and the turning threshold $\alpha_{c,1}$:
\begin{align}
    &W_{c,1} = d_{c,1} \, \mathtt{rad2vec}( \beta_c ) + \mathtt{proj}_{\positions}(\state_{\mathrm{ism},0})\\
    & \quad \text{where} \notag \\
    &  \quad\beta_c  = \begin{cases}
        \mathtt{proj}_{\orientation} (\state_{\mathrm{ism},0}) - \alpha_\mathrm{c,1} & \text{if } \varphi_{c,1},\\
        \mathtt{vec2rad}(\mathtt{proj}_{\positions} (\state_{\mathrm{obs},0}), \mathtt{proj}_{\positions} (\state_{\mathrm{ism},0})) & \text{else};
    \end{cases} \notag \\
    & \quad \varphi_{c,1}: \mathtt{vec2rad}(\mathtt{proj}_{\positions} (\state_{\mathrm{obs},0}), \mathtt{proj}_{\positions} (\state_{\mathrm{ism},0})) \notag \\
    &
    \quad\quad\quad\quad\quad\leq \mathtt{proj}_{\orientation} (\state_{\mathrm{ism},0}) - \alpha_\mathrm{c,1}. \notag
\end{align}
Intuitively, the two cases are: if the position of the other vessel $\mathtt{proj}_{\positions} (\state_{\mathrm{obs},0})$ is at an angle less than $\alpha_{c,1}$ to the right of the \ac{ism} vessel, then the waypoint $W_{c,1}$ would be generated on the line $\alpha_{c,1}$ right of the \ac{ism} vessel. Otherwise, the line is generated by connecting the positions of the vessels. On defined line, the waypoint $W_{c,1}$ is positioned at a distance $d_{c,1}$ (see Fig.~\ref{fig:waypointadaption}(b)).  
The next waypoint $W_{c,2}$ is a guiding waypoint and is generated in the direction that is perpendicular to the orientation of the vessel when the crossing maneuver starts, i.e., $\mathtt{proj}_{\orientation} (\state_{\mathrm{ism},0}) - \pi/2$. The \ac{ism} vessel tracks the waypoint $W_{c,2}$ until the other vessel is least $d_{c,2}$ behind it, i.e., $\mathrm{in\_h}(\cdot, d_{c,2})$ is true, and $\mathrm{stable\_orientation}$ is true for the orientation specified by the positions $W_{c,1}$ and $W_{c,2}$, i.e., at time point $t_2$. 
Then, the guiding waypoint $W_{c,3}$ is generated to direct the \ac{ism} vessel along a path parallel to its original orientation when the encounter started, i.e., is created in the direction $\mathtt{proj}_{\orientation} (\state_{\mathrm{ism},0})$. Once the other vessel is behind the \ac{ism} vessel by at least the distance $d_{c,3}$, i.e., $\mathrm{in\_h}(\cdot, d_{c,3})$ is true, and the predicate $\mathrm{orientation\_stable}$ is true for the orientation specified by the position of the vessel at $t_2$ and waypoint $W_{c,3}$, the crossing collision avoidance maneuver is terminated and the vessel resumes to its original planning problem.

\paragraph{Overtake}
The overtaking traffic rule specifies that the slower vessel can be overtaken on any side, while the orientation of the overtaking vessel needs to change significantly. Thus, we first determine the overtaking side based on the relative angle between the two vessels. If the orientation of the other vessel is the same or more to the left than the orientation of the \ac{ism} vessel, the \ac{ism} vessel will overtake on the right side and otherwise on the left. This logic for determining the side results in an increasing or stable distance between the two vessels if the other vessel keeps its orientation.
The first and normal waypoint $W_{o,1}$ is positioned perpendicular to the orientation of the other vessel at a distance that is at least $d_{o,1}$ and so that the orientation for the \ac{ism} vessel has to change at least $\alpha_{o,1}$ (see Fig.~\ref{fig:waypointadaption}(c)). To this end, we first compute the intersection point $\positions^{o}$ between the line perpendicular to the other vessel and the minimal turned orientation of the  \ac{ism} vessel:
\begin{align}
    \positions^{o} &= \mathtt{intersect}(g_2, g_1) \, \text{where} \\
    &g_1 = \langle \mathtt{proj}_{\positions} (\state_{\mathrm{ism},0}), \mathtt{proj}_{\orientation} (\state_{\mathrm{ism},0}) \pm \alpha_{o,1} \rangle, \notag \\
    &g_2 = \langle\mathtt{proj}_{\positions} (\state_{\mathrm{obs},0}), \mathtt{proj}_{\orientation} (\state_{\mathrm{obs},0}) \pm \pi/2   \rangle.\notag 
\end{align}
Note that the $+$ case is when left is the overtaking side and $-$ when right is the overtaking side (e.g., as in Fig.~\ref{fig:waypointadaption}(c)).
\begin{align}
    &W_{o,1} = d_{o}^{\prime} \, \mathtt{rad2vec}( \beta_o ) + \mathtt{proj}_{\positions}(\state_{\mathrm{ism},0})\\
    & \quad \text{where} \notag \\
    &  \quad\beta_o  = \begin{cases}
        \mathtt{proj}_{\orientation} (\state_{\mathrm{ism},0}) \pm \alpha_\mathrm{o,1} & \text{if } \varphi_{o,1},\\
        \mathtt{vec2rad}(\positions_o, \mathtt{proj}_{\positions} (\state_{\mathrm{ism},0})) & \text{else};
    \end{cases} \notag \\
    & \quad \varphi_{o,1}: \mathtt{vec2rad}(\positions_o, \mathtt{proj}_{\positions} (\state_{\mathrm{ism},0})) \notag \\
    &
    \quad\quad\quad\quad\quad\leq \mathtt{proj}_{\orientation} (\state_{\mathrm{ism},0}) - \alpha_\mathrm{o,1}, \notag
\end{align}
where $d_{o}^{\prime}$ is calculated so that $W_{o,1}$ is on line $g_2$ (see Fig.~\ref{fig:waypointadaption}(c)).
The second and guiding waypoint $W_{o,2}$ is positioned so that the direction from $W_{o,1}$ to $W_{o,2}$ is parallel to the orientation of the \ac{ism} vessel when the encounter situation started, i.e., $\mathtt{proj}_{\orientation} (\state_{\mathrm{ism},0})$.
The overtaking avoidance maneuver is terminated once the other vessel is behind the \ac{ism} vessel with at least the distance $d_{o,2}$, i.e., $\mathrm{in\_h}(\cdot, d_{o,2})$ is true, and the orientation of the vessel is stable with respect to the orientation defined by the waypoints $W_{o,1}$ and $W_{o,2}$.

\paragraph{Resuming to specified motion planning problem}\label{paragraph:resume2path}
Once the encounter situation terminates, the \ac{ism} vessel has to continue on the path characterized by the waypoints $\mathcal{G}$ to fulfill the motion planning task. Thus, the waypoint list is appended with all the waypoints from $\mathcal{G}$ that have not been reached already.

\subsection{\ac{mpc}-based low-level control}\label{sc:opimitationmpcformulation}
To obtain the objective for the optimal control formulation, we compute a trajectory of desired positions based on the current position of the \ac{ism} vessel, the last reached waypoint, and the list of waypoints to be reached next. In particular, we connect the waypoints with the next one with line segments starting at the last reached waypoint. Then, we perpendicularly project the current position of the vessel $\positions_0$ onto the line segment of the last reached $W_{j-1}$ and next waypoint $W_j$:
\begin{equation}
    \positions^p = \mathtt{orthop}(\positions_0, W_{j} - W_{j-1})
\end{equation}
The projected position $\positions^p$ is the desired position at the current time $\desiredpositions_0$. The trajectory of desired positions is computed by progressing along the line segments with $\nvelocity^d \, \Delta t \, k$ where $k \in \{1,...,N\}$. Note that $\nvelocity^d$ is the desired velocity of the vessel and the generation of the desired positions is denoted by the function $\mathrm{get\_reference}(\mathcal{W}, \state)$. 

With the desired positions, we formulate a model predictive controller that tracks the positions. In particular, we solve the optimization problem:
\begin{subequations}
\begin{align}
    \control^{\ast}_{\{1,...,T\}} &= \underset{\control_{\{1,...,T\}}}{\arg}\min \, \sum_{t=0}^{T} \|\positions_{t+1} - \desiredpositions_{t+1} \|_2  \label{eq:MPC}\\
    \text{subject to} & \, \,  \state_{t+1} = g(\state_{t}, \control_{t+1}) \, \forall t \in \{0, ..., T-1\} \\
    &\;\control_{t} \in \mathcal{U} \; \forall t \in \{1, ..., T\} \\
    &\;\;\state_0 \in \mathcal{S}.
\end{align}
\end{subequations}
We linearize the continuous vessel dynamics of Eq.~(\ref{eq:yawconstrainedmodel}) with a first-order Taylor approximation at the current state $\state_0$ and at the current control signal $\control_0$. With the approximated linear dynamics we can obtain the discrete-time dynamics $g(\state,\control)$. Thus, the overall optimization problem is a quadratic program with affine constraints for the dynamics and the state and input constraints, which can be efficiently solved. 

\begin{algorithm}[tb]
\caption{\ac{ism} algorithm}
\label{alg:ism}
\begin{algorithmic}[1]
\State \textbf{Input:} waypoints $\mathcal{W}$, vessel parameters $\Gamma$, environment $E$, initial state $\state$, maximum simulation steps $t_{\mathrm{max}}$
\State $t \gets 0$
\While{$\lnot$ termination}
    \State $\mathcal{W} \gets \text{update\_waypoints}(\mathcal{W}, \state, E)$
    \State $\xi^d \gets \text{get\_reference}(\mathcal{W}, \state)$
    \State $\control_0 \gets \text{MPC}(\xi^d, s, \Gamma)$
    \State $(\state, E) \gets \text{step\_environment}(u)$
    \State $t \gets t + 1$
    \If{goal reached or $t = t_{\mathrm{max}}$ or collision}
        \State termination $\gets$ True
    \EndIf
\EndWhile
\end{algorithmic}
\end{algorithm}

The overall \ac{ism} algorithm is composed of the \waypointadaption{} and the model-predictive controller, which are used to react according to the traffic situation at each time step (see Algorithm~\ref{alg:ism}). First, the clock is initialized. Then, at every time step, the \waypointadaption{} is called to update the waypoint list $\mathcal{W}$, desired positions $\positions^d$ and the control signal $\control$ are calculated, and the \ac{ism} vessel and environment is simulated for a time step. This process continues until the goal waypoint is reached, the maximum simulation time $t_\mathrm{max}$ is reached, or a collision occurs. Note that our \ac{ism} algorithm is a discrete-time model, which is commonly not a limitation for simulation settings.

\section{Numerical Experiments}\label{sec:experiments}

We investigate a diverse and challenging set of maritime traffic situations, ranging from an extensive amount of critical reactive traffic to scenarios based on real-world traffic data, to evaluate the coherence, effectiveness, modularity, and scalability of our \ac{ism}. To allow for reproducibility, we specify the traffic situations as CommonOcean \cite{Krasowski2022.CommonOcean} benchmarks\footnote{The benchmarks will be published with the final version and the implementation is available at \href{https://commonocean.cps.cit.tum.de/commonocean-sim}{commonocean.cps.cit.tum.de/commonocean-sim}.}. 
CommonOcean benchmarks define the motion planning problems for the \ac{ism} vessels by an initial state and goal state, and optionally contain trajectories of other vessels. In particular, we regard traffic situations where all vessels are simulated with our proposed \ac{ism} and mixed traffic situations with one \ac{ism} vessel and one dynamic obstacle that is non-reactive. For all \ac{ism} vessels in our experiments, the waypoint list consists of one goal waypoint, and we use a time step size of $\SI{1}{\second}$, which is appropriate for maritime navigation. We consider five evaluation settings:
\begin{enumerate}[label=\alph*)]
    \item three illustrative encounter situations for crossing, head-on, and overtaking,
    \item 2,000 safety-critical situations with two \ac{ism} vessels,
    \item 2,000 safety-critical situations with one dynamic obstacle and an \ac{ism} vessel in each,
    \item 49 critical situations reconstructed from \ac{ais} data with one \ac{ism} vessel,
    \item a complex multi-vessel traffic situation with six \ac{ism} vessels. 
\end{enumerate}
Note that the traffic situation sets b), c), and d) are based on the scenarios used for evaluation in \cite{Krasowski2024.safeRLautonomousVessels}. Specifically, c) and d) are the same benchmarks used in \cite{Krasowski2024.safeRLautonomousVessels}, and b) replaces the dynamic obstacle with an \ac{ism} vessel where the initial state is the initial state of the dynamic obstacle and the goal waypoint is the final state of the dynamic obstacle. 
We use two-vessel traffic situations for the large-scale evaluations in b) to d) since these are the most common, and if only interactive \ac{ism} vessels are present as in b), there should be no collisions despite the criticality.
Two vessels are employed for our experiments: a container ship (type 1) and a tanker (type~2). These are CommonOcean vessels 1 and 2, which are the most relevant for open sea traffic\footnote{\href{https://gitlab.lrz.de/tum-cps/commonocean-vessel-models}{gitlab.lrz.de/tum-cps/commonocean-vessel-models}}. Their parameters are based on the SR 108 container ship and the lake freighter simulation provided in the Marine Systems Simulator\footnote{\href{https://github.com/cybergalactic/MSS}{github.com/cybergalactic/MSS}}. We choose a high desired velocity for the tanker to obtain behavior similar to the container ship and use the same time horizons and parameters for waypoint adaption.
Since the evaluation for scenario sets a) and e) are mainly illustrative, we only show results for the container ship. We specify all parameters for the \ac{ism} in Table~\ref{tab:vesselparameters}. To ensure a sufficient distance to guiding waypoints, we generate them $10^6$~\si{\meter} ahead.

To evaluate the high-level, \waypointadaption{} of the \ac{ism}, we evaluate the traffic rule compliance with respect to the formalized traffic rules in \cite{Krasowski2024.safeRLautonomousVessels}. We use the same parameters for the rules, except for $t_\mathrm{maneuver}$, which we increase to \SI{90}{\second}. The performance of the low-level \ac{mpc} controller is measured by the deviation from the reference positions, and the average absolute control inputs for turning rate $\omega$ and normal acceleration $\nacceleration$. We run all experiments on a machine with an AMD EPYC 7742 2.2 GHz
processor and 1024 GB of DDR4 3200 MHz memory and use Gurobi as solver for the optimization problem in Eq.~\eqref{eq:MPC}.

\begin{table}
    \renewcommand{\arraystretch}{1.3}
    \caption{Parameters for numerical experiments}
    \label{tab:vesselparameters}
    \centering
    \begin{tabular}{c l l l l}
    \toprule
    \textbf{Variable} & \textbf{Description} & \multicolumn{2}{l}{\textbf{Value}} & \textbf{Unit} \\ \midrule
    \multicolumn{5}{l}{\textit{Vessel parameters}}  \\ \midrule
    & & Type 1 & Type 2 & \\\cmidrule{3-4}
         $v_\mathrm{max}$ & maximum velocity    & $16.8$ & $7.02$ &\unit{\meter\per\second}\\
         $v_\mathrm{des}$ & desired velocity & $8.4$& $7.02$ &\unit{\meter\per\second}\\
         $\omega_\mathrm{max}$ & maximum yaw  & $0.03$ & $0.0078$& \unit{\radian\per\second}\\
         $a_\mathrm{max}$ & maximum acceleration  &   $0.24$ &   $0.0127$ &\unit{\meter\per\second\squared}\\
         $l$ & vessel length & $175$& $304.8$&\unit{\meter} \\
         $w$ & vessel width & $25.4$ & $32$&\unit{\meter} \\
        \midrule 
         \multicolumn{5}{l}{\textit{Waypoint adaption parameters}}  \\ \midrule
        $\Delta t$ & time step & \multicolumn{2}{l}{$1.0$} &\unit{\second}\\
        $\alpha_{h,1}$  & head-on turning angle  & \multicolumn{2}{l}{$0.8$} & \unit{\radian}\\
        $d_{h,1}$  &  head-on turning distance & \multicolumn{2}{l}{$l+w$} &\unit{\meter}\\
        $d_{h,2}$ &  head-on parallel distance  & \multicolumn{2}{l}{$2l$} &\unit{\meter}\\
        $\alpha_{c,1}$  &  crossing turning angle &  \multicolumn{2}{l}{$0.785$} &\unit{\radian} \\
        $d_{c,1}$  &  crossing turning distance & \multicolumn{2}{l}{$1.5 \, \frac{\alpha_{c,1} \, v_\mathrm{des}}{\turningrate_\mathrm{max}}$} & \unit{\meter}\\
        $d_{c,2}$  &  crossing parallel distance  & \multicolumn{2}{l}{$2l$} &\unit{\meter} \\
        $d_{c,3}$  &  crossing rear distance & \multicolumn{2}{l}{$2l + 2w$} &\unit{\meter} \\
        $\alpha_{o,1}$  &  overtaking turning angle & \multicolumn{2}{l}{$0.261$} & \unit{\radian} \\
        $d_{o,1}$  &   overtaking obstacle distance & \multicolumn{2}{l}{$2l + 2w$} & \unit{\meter} \\
        $d_{o,2}$  &  overtaking parallel distance  & \multicolumn{2}{l}{$2l$} &\unit{\meter}\\
        $t_{so}$  &  stable orientation time  & \multicolumn{2}{l}{$10 \Delta t$} & \unit{\second} \\
        $\alpha_{so}$  & angle for stable orientation  & \multicolumn{2}{l}{$0.005$} &\unit{\radian} \\
        \midrule
         \multicolumn{5}{l}{\textit{\ac{mpc} parameters}}  \\ \midrule
        $T$ & \ac{mpc} prediction horizon & \multicolumn{2}{l}{$90$} & \unit{\second} \\
        $d_{wp}$ & waypoint reaching distance & \multicolumn{2}{l}{$0.5 l$} & \unit{\meter}\\
        $d_\mathrm{term}$ & goal reaching distance & \multicolumn{2}{l}{$0.25 l$} & \unit{\meter}\\
         \bottomrule
    \end{tabular}
\end{table}

\subsection{Results}

\begin{figure*}[ht]
 \centering
\begin{tabular}{lll} 
	\centering
	
	\begin{subfigure}[b]{5.55cm}
		\includegraphics{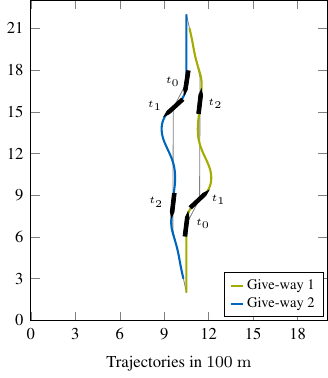}
	\end{subfigure} \hfill
	&  
	\begin{subfigure}[b]{5.55cm}
		\includegraphics{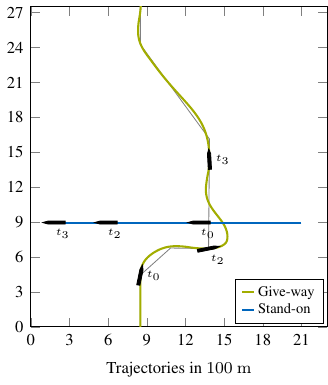}
	\end{subfigure}   \hfill
	& 
	\begin{subfigure}[b]{5.55cm}
		\includegraphics{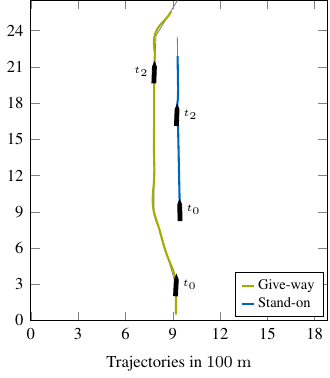}
	\end{subfigure}  
	
	\\
	
	\begin{subfigure}[t]{4.0cm}
		\includegraphics{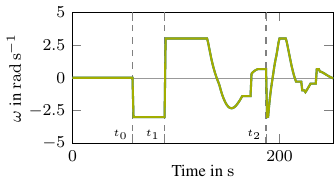}
	\end{subfigure} 
	&  \hspace{0.04cm}
	\begin{subfigure}[t]{4.0cm}
		\includegraphics{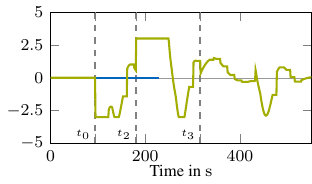}
	\end{subfigure}   
	&  \hspace{0.04cm} 
	\begin{subfigure}[t]{4.0cm}
		\includegraphics{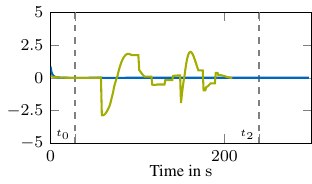}
	\end{subfigure}  
	
	\\
	
	\begin{subfigure}[t]{4.0cm}
		\includegraphics{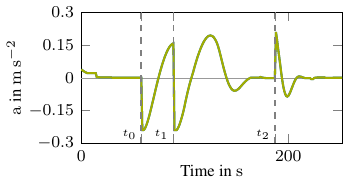}
		
	\end{subfigure} 
	& 
	\begin{subfigure}[t]{4.0cm}
		\includegraphics{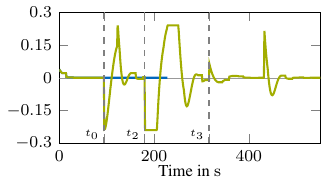}
	\end{subfigure}   
	&  
	\begin{subfigure}[t]{4.0cm}
		\includegraphics{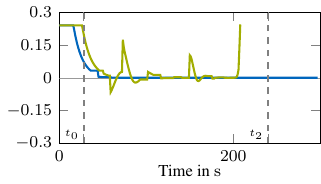}
	\end{subfigure}  \\
	\hspace{2.5cm} (a) head-on & \hspace{2.5cm} (b) crossing & \hspace{2.5cm} (c) overtake
\end{tabular}
     
        \caption{Trajectories and control inputs for three encounter situations: head-on, crossing, and overtaking (left to right). The marked time steps are as specified in Fig.~\ref{fig:waypointadaption}. The upper row displays the trajectories in the position space with the optimal paths described by the waypoints in gray. The vessel is of type 1 and is depicted with its actual spatial dimension. The second and third rows show the angular velocity and normal acceleration over time, respectively.}
        \label{fig:illustative-results-plots}
\end{figure*}

\paragraph{Illustrative encounter situations}
First, we investigate the effectiveness of the \ac{ism} on three illustrative situations reflecting three situations where the \waypointadaption{} generates \adaptivewaypoints{} as conceptually described in Fig.~\ref{fig:waypointadaption}. Fig.~\ref{fig:illustative-results-plots} displays the observed trajectories and control input. The \ac{ism} vessels track their \adaptivewaypoints{} closely except when sharp turns are required, e.g., at $t_1$ in the head-on situation. The reason for the deviation is that the vessel dynamics do not allow for too sharp turns. In particular, we observe in the control inputs that give-way vessels are slowing down and turning after time steps with a waypoint list adaption, e.g., $t_1$ in head-on. Note that for the head-on encounter, the two vessels produce exactly the same control input. If the \ac{ism} vessel is a stand-on vessel, the control input is zero since it has to keep its course and speed. Overall, Fig.~\ref{fig:illustative-results-plots} shows a coherent and rule-compliant behavior of \ac{ism} vessels in the three encounter situations specified on the \ac{colregs}.

\begin{table*}[tb]
\renewcommand{\arraystretch}{1.3}
\caption{Evaluation of \ac{ism} with different vessel types for safety-critical traffic situations}
\label{tab:largescaleresults}
\centering
\begin{tabular}{l@{\extracolsep{10pt}} c c@{\extracolsep{10pt}} c c  @{\extracolsep{10pt}} c c } 
\toprule
  & \multicolumn{2}{c}{\ac{ism} only} & \multicolumn{2}{c}{Mixed traffic} & \multicolumn{2}{c}{\ac{ais} scenarios}\\ \cmidrule{2-3} \cmidrule{4-5} \cmidrule{6-7}
  & Container ship & Tanker & Container ship & Tanker & Container ship & Tanker\\
\midrule 
  Goal reached in \% & 98.2 & 98.3 & 97.0  & 95.6 & 98.0 & 95.9\\
  Collisions in \% & 0.0 & 0.0 & 2.8  & 4.1  & 0.0 & 0.0 \\ \midrule
  $R_3$ in \% & 99.5 & 89.1 & 98.9  & 96.2  & 100 & 100 \\
  $R_4$ in \% & 99.7 &  99.5 & 99.5 & 99.7  & 100 & 100 \\
  $R_5$ in \% & 98.2 & 100 & 99.9  & 100 & 100 & 100 \\
  $R_6$ in \% & 96.3 & 94.2 & 96.9  & 95.3 & 100 & 100 \\
  $\bigwedge_{i \in I} R_i$ in \% &  94.9  & 83.4 & 95.4 & 91.3 & 100 & 100 \\ \midrule
$\| \positions - \positions^d\|_2$ in \unit{\meter} & $4.510 \pm 8.298$ & $61.71 \pm 95.04$ & $2.955 \pm 8.200$  & $34.60 \pm 85.94$  & $1.494 \pm 5.382$ & $32.06 \pm 95.55$  \\
$\nacceleration$ in \unit{\meter\per\second\squared} & $0.012 \pm 0.016$ & $0.004 \pm 0.004$ & $0.010 \pm 0.014$  & $0.003 \pm 0.003$  & $0.007 \pm 0.011$ & $0.002 \pm 0.003$ \\
$\turningrate$ in \unit{\radian\per\second}& $0.003 \pm 0.004$ & $0.002 \pm 0.002$ & $0.002 \pm 0.003$  & $0.0018 \pm 0.002$  & $0.001 \pm 0.002$ & $0.001 \pm 0.002$\\
\bottomrule
\end{tabular}

\vspace{0.3cm}
{\scriptsize Note: The traffic rules are abbreviated with $R_i$, where $i=3$ is the give-way rule for crossing, $i=4$ for head-on, $i=5$ for overtaking, and $i=6$ is the rule for stand-on vessels. The subscripts are consistent with \cite{Krasowski2024.safeRLautonomousVessels}, where also the formalization is detailed.}
\end{table*}    

\paragraph{\ac{ism} only}
To test the \ac{ism} at scale, we evaluate it on 2000 safety-critical traffic situations where the shortest paths between initial states and goal waypoints of the vessels are crossing each other. The results are presented in Table~\ref{tab:largescaleresults}. The \ac{ism} vessels never collide and reach the goal in about $98 \%$. The traffic rule compliance of the sailed trajectories is $95 \%$ for the container ship and $85 \%$ for the tanker. Note that this lower traffic rule compliance for the tanker is mainly due to the lower compliance with the give-way crossing rule $R_3$. We investigated this in detail and observed that the tanker would require a longer maneuver time $t_\mathrm{maneuver}$ due to its lower control input bounds. If we extend the maneuver time by $\SI{15}{\second}$, the rule compliance with $R_3$ increases to $98 \%$ and the conjunction of all rules to $92 \%$. 
For the control metrics, i.e., position deviation, acceleration, and yaw, the low-level \ac{mpc} can keep the container ship closer to the desired path than the tanker, which is due to its better maneuverability. The similar average for yaw for the two vessel types is due to the vessels maneuvering straight for a significant part of their trajectories. Yet, the higher standard deviation and higher average acceleration for the container ship arise from its better maneuverability. 
Generally, the results highlight the effectiveness of the \ac{ism} with its high goal-reaching rate and rule-compliant collision avoidance and the transferability between different vessel types. 
Lastly, we conduct a brief sensitivity analysis for the tanker with respect to the crossing parallel distance $d_{c,2}$ by setting it to $3l$ and $5l$. Increasing $d_{c,2}$ leads to less goal reaching, $98.2 \%$ and $97.4 \%$, respectively, and lower average position deviation, $61.59 \unit{\meter}$ and $59.01 \unit{\meter}$, while the rule compliance percentages do not change.

\paragraph{Mixed traffic}
To show that the performance of the \ac{ism} is competitive with other autonomous vessel navigation algorithms and effective in mixed traffic situations where only some vessels are controlled by the \ac{ism}, we use the 2,000 benchmarks with one dynamic obstacle from \cite{Krasowski2024.safeRLautonomousVessels}. The results are shown in columns three and four of Table~\ref{tab:largescaleresults}. For these traffic situations, the \ac{ism} vessels achieve a high goal-reaching rate, while about $3 - 4 \%$ of the 2,000 traffic scenarios lead to a collision, depending on the vessel type. These collisions occur when the dynamic obstacle is not reacting according to traffic rules, which underscores the need for interactive vessel models like the \ac{ism} to realistically benchmark autonomous vessels. The overall traffic rule compliance is very high, with $96 \%$ for the container ship and $92 \%$ for the tanker. Note that the goal-reaching performance is higher than for the reinforcement learning agents trained in \cite{Krasowski2024.safeRLautonomousVessels}, where the highest goal-reaching rate on a test set of 400 of the 2,000 scenarios is $90.7 \%$. At the same time, the collision rate is comparable (about $3 \%$ is reported in \cite{Krasowski2024.safeRLautonomousVessels} for the container ship). For the control signals and deviation from the desired positions in Table~\ref{tab:largescaleresults}, we observe a similar pattern as for the \ac{ism} only results.

\paragraph{\ac{ais} scenarios}
Handcrafted critical scenarios may lack the diversity encountered in real traffic. To also evaluate our \ac{ism} with more realistic dynamic obstacles, we use mixed traffic situations based on \ac{ais} data of maritime traffic. \ac{ais} is a radio safety system that virtually all seagoing vessels are equipped with. \ac{ais} data contains static information about a vessel, e.g., its length, as well as dynamic information, e.g., its current speed and location. The 49 critical scenarios, where the paths of the recorded vessels intersect and the vessels are in close proximity, are filtered from the Marine Cadastre dataset \cite{MarineCadastre.2009} for US coastal regions in January 2019 (see \cite{Krasowski2021.MarineTrafficRules, Krasowski2024.safeRLautonomousVessels} for a detailed description of the scenario preprocessing and filtering). The results are reported in Table~\ref{tab:largescaleresults}. We again observe high goal-reaching rates above $95 \%$ and no collisions. Additionally, the vessels always comply with the traffic rules.  Generally, there are fewer avoidance maneuvers necessary for this set of scenarios, leading to lower averages for path deviation and control input. Note that the statistical power of this setup is lower due to a significantly lower number of scenarios than for \ac{ism} only and mixed traffic. However, increasing the statistical power is not trivial since these 49 scenarios have been filtered from about 30 GB of US-coastal traffic data of January 2019 \cite{Krasowski2024.safeRLautonomousVessels}.

\begin{figure}
    \centering
     \includegraphics{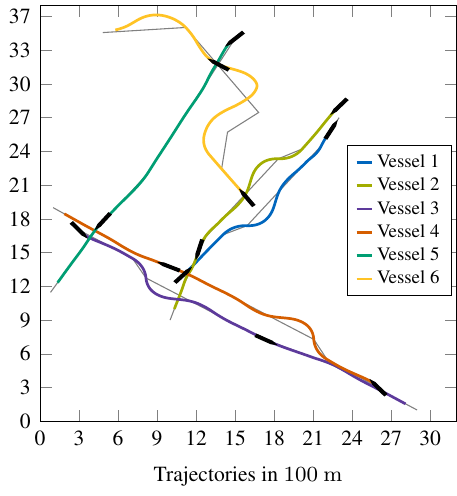} 
    \caption{Multi \ac{ism} vessel scenario with colored position trajectories and gray optimal paths for six vessels. The vessels are displayed in black with their proportional spatial dimensions at the initial time step and the time point \SI{236}{\second}.}
    \label{fig:busy-scenario}
\end{figure}

\paragraph{Multi-vessel traffic situation}
The \ac{colregs} specify collision avoidance between two vessels. While two-vessel encounters are prevalent on the open sea, Fig.~\ref{fig:busy-scenario} demonstrates the capabilities of the \ac{ism} for a traffic situation with six vessels. The vessels perform collision avoidance maneuvers with respect to the vessels for which a conflict is first detected. In particular, we observe head-on collision avoidance between vessels 1 and 2, and between vessels 3 and 4, while vessels 5 and 6 are in a crossing situation. To demonstrate the scalability of the \ac{ism}, we empirically evaluate the computational runtime when iteratively increasing the number of traffic participants. For one vessel, we observe an average runtime per time step of \SI{0.0653}{\second} while on average  \SI{0.0649}{\second} are attributed to solving the \ac{mpc} problem. The average runtime per step increases approximately linear with the number of vessels to \SI{0.336}{\second} on average for six vessels while \SI{0.328}{\second} and \SI{0.0057}{\second} are originating from the \ac{mpc} problem and waypoint adaption, respectively. This is still significantly below the time step size of \SI{1}{\second}. Note that we did not parallelize solving the \ac{mpc} problem for the different vessels, and our implementation is in Python. Changing the coding language and using parallelization would likely significantly reduce the reported runtime.  

\section{Discussion and Limitations}\label{sec:discussion}

\textbf{Traffic rule violations and emergency rule. }
Our experimental evaluation shows that the proposed \ac{ism} performs well in different traffic situations and for different vessel types. Although we use similar predicates in the \waypointadaption{} and in the formalized traffic rules \cite[Table~I]{Krasowski2024.safeRLautonomousVessels}, we observe some violations of the formalized traffic rules. The main cause for the rule violations is that the time horizon specified to reach a state in which no collision is possible is sometimes not long enough. Thus, the rule violations are a feature of evaluating using unambiguously formalized traffic rules.
We do not observe collisions in the \ac{ism} only setting, which indicates that we can simulate collision-free traffic with the \ac{ism}. For the mixed traffic setting, the collisions are due to the non-\ac{ism} vessel not appropriately reacting.
Thus, to improve the usability of the \ac{ism} in mixed traffic settings, future work should target adding an emergency mode that is activated if two vessels are about to collide. While there are some existing works that could be integrated as a controller for the emergency mode \cite{koszelew2017determination, Krasowski2024.safeRLautonomousVessels}, in traffic with more than two vessels, it is challenging to define a computationally efficient controller that is capable of handling all possible multi-agent conflict configurations.

\textbf{Evaluation scope and implementation efficiency. }
Our experiments consider six different setups and our 2,000 scenarios for the \ac{ism} only and mixed traffic setting extensively evaluate the proposed \ac{ism}. However, there are still many more settings that could be considered, e.g., using the \ac{ism} with other controllers or different parameters, real-world experiments, or large-scale evaluations with \ac{ais} data. For broad use of the \ac{ism}, the runtime of our implementation could be improved by parallelizing the \ac{mpc} computations. Additionally, we recompute the control input at every time step. However, for navigation sections where the vessel mainly sails straight, we empirically observed that using the previous \ac{mpc} result does not significantly impact the performance. Thus, we could further improve the runtime by using previously computed \ac{mpc} results if the deviation between the predicted and the observed state is small. We added this runtime improvement feature in CommonOcean-Sim and believe that this open-source software package will facilitate future research with more than the presented traffic settings.

\textbf{Vessel dynamics fidelity. }
The collision avoidance path described through the waypoints contains sudden changes of direction, which are also present in the generated reference positions. The maneuverability of the vessels usually does not allow for too sharp turns, so we observe significant deviations from the path. This also leads to challenges for the \ac{mpc} because the computed cost might be smaller when accelerating to track the desired positions before a turn well than slowing down and tracking earlier states with more error to reduce the error after a turn.
As a result, the vessel's tracking of the path after a turn is significantly worse than necessary. This issue could be mitigated by smoothing the desired positions using methods as in \cite{Wuersching2024}. Additionally, the yaw-constrained model (see \eqref{eq:yawconstrainedmodel}) captures the turning capabilities of a vessel, but it still simplifies the real-world dynamics significantly. Thus, future work should investigate \ac{mpc} for higher-fidelity models, e.g., the three-degrees-of-freedom model~\cite{Fossen2011}. Using a model with higher fidelity has the additional benefit that integrating disturbances originating from wind and waves can be considered at the cost of increased computation runtime.

\textbf{Traffic rule formalization. }
Our \waypointadaption{} is effective and can be easily adapted due to its parametrization. The validity is supported by the fact that we do not observe collisions in the \ac{ism} only setting, and in the mixed traffic setting, the \ac{ism} vessels are always stand-on vessels when colliding. Our \waypointadaption{} is built on the predicates defined in \cite{Krasowski2021.MarineTrafficRules}. However, there are also other metrics \cite{Heiberg2022} and formalizations \cite{Torben2023} that could be used to model \ac{colregs} rules. Additionally, the current version of the \ac{ism} does not capture collision avoidance behavior between other vessel types, e.g., a sailing vessel has priority over a power-driven vessel. Thus, future work should investigate other traffic rule metrics and extensions to integrate more vessel types and traffic situations.

\section{Conclusion}\label{sec:conclusion}
This paper introduces the intelligent sailing model (ISM) that models collision avoidance behavior of vessels on the open sea while solving a motion planning problem. The \ac{ism} architecture consists of a high-level \waypointadaption{} and a low-level \ac{mpc} that tracks the waypoints and allows for changing vessel types. We perform an extensive numerical evaluation showing that the \ac{ism} solves motion planning problems effectively with high rule compliance in mixed traffic and \ac{ism} only settings. The \ac{ism} is competitive with other autonomous vessel navigation algorithms in both hand-crafted and recorded real-world traffic scenarios, and it scales to high traffic density. The parametrization of the \ac{ism} makes adaptions easy and interpretable. We believe that the \ac{ism} will lead to a more realistic and rigorous evaluation of motion planning research for vessels and accelerate the development of autonomous navigation on the open sea.

\section*{Acknowledgment}

The authors gratefully acknowledge the partial financial support of this work by Deutsche
Forschungsgemeinschaft (DFG, German Research Foundation) - SFB 1608 - under grant 501798263 and the National Science Foundation grant {CNS-2111688}. 

\printbibliography{}

\end{document}